\pdfoutput=1

\documentclass[11pt]{article}

\usepackage[final]{acl}
\usepackage{multirow}
\usepackage{xcolor}
\usepackage{color, colortbl}

\usepackage{pifont}
\newcommand{\cmark}{\ding{51}}
\newcommand{\xmark}{\ding{55}}

\definecolor{mygray}{gray}{0.9}
\usepackage{booktabs}
\usepackage{amsmath} 
\usepackage{amssymb}
\usepackage{listings}

\usepackage{algorithm}
\usepackage{algpseudocode}
\usepackage{times}
\usepackage{latexsym}

\usepackage[T1]{fontenc}

\usepackage[utf8]{inputenc}

\usepackage{microtype}

\usepackage{inconsolata}
\usepackage{subcaption}  

\usepackage{graphicx}

\title{What Makes Good Instruction-Tuning Data? An In-Context Learning Perspective}
\author{
Guangzeng Han \\
University of Memphis \\
\texttt{ghan@memphis.edu}
\And
Xiaolei Huang \\
University of Memphis \\
\texttt{xiaolei.huang@memphis.edu}
}

\usepackage{bibentry}

\begin{document}

\maketitle
\begin{abstract}
Instruction-tuning datasets often contain substantial redundancy and low-quality samples, necessitating effective data selection methods. 
We propose an instruction data selection framework based on weighted in-context influence (wICI), which measures how effectively each candidate example reduces instruction-following difficulty for semantically related peers. 
Through systematic experiments, we address three key questions: what constitutes effective instruction tuning data from an in-context perspective, whether sample difficulty correlates with in-context influence, and how in-context influence translates to instruction tuning effectiveness.
Experiments across multiple models and benchmarks demonstrate that our method consistently outperforms existing baselines under constrained data budgets, while empirically showing that sample difficulty negatively correlates with in-context influence.\footnote{The code is available at: \url{https://github.com/trust-nlp/SyntheticData-Curator}}
\end{abstract}

\section{Introduction}

In-context learning (ICL)~\citep{dong2024survey} enables large language models (LLMs) to adapt its behavior at inference time by conditioning on a small set of demonstration examples. 
By prepending a handful of instruction–response pairs to the input prompt, the model leverages its pre-trained knowledge to generalize to new tasks without any parameter updates. 
Previous work has explored how to select effective demonstrations, using criteria such as semantic similarity~\citep{li2023finding,dong2024survey}, diversity, or model feedback~\citep{wang2024learning,ye2023compositional}, to maximize performance on a certain task.
Instruction tuning~\citep{lou2024large}, by contrast, updates model parameters through fine-tuning on large collections of instruction–response pairs~\citep{wang2023self-instruct,xu2023wizardlmempoweringlargelanguage}. 
This paradigm has proven effective at improving instruction-following ability, but assembling high-quality tuning datasets is costly. 
Data selection methods like Superfiltering~\citep{li2024superfiltering} use model perplexity to prune simple examples, and DEITA~\citep{liu2024what} apply learned reward models to rank and filter instruction–response pairs based on human-aligned quality signals.

Despite the successes of each paradigm in isolation, the relationship between examples that perform well in in-context learning and those that constitute valuable instruction-tuning data remains underexplored.
Recognizing this gap, ~\citep{li2024oneshot} made a pioneering attempt in NUGGETS to bridge in-context learning and instruction tuning data curation paradigms. 
NUGGETS evaluates data quality by using each candidate instruction as a one-shot demonstration and measuring its impact on performance across a fixed global anchor set, establishing the first connection between ICL and instruction tuning.
While NUGGETS made significant progress in this direction, we explore how this paradigm can be further advanced. Specifically, NUGGETS evaluates all candidates using the same fixed global anchor set regardless of semantic relevance, employs simple binary scoring without considering improvement magnitude or task difficulty, and requires substantial computational resources by evaluating each candidate on large anchor sets (typically 1,000 samples).
We hypothesize that by introducing dynamic, semantically relevant probe sets and generalization-based weighting mechanisms, we can more precisely measure the influence of samples while significantly improving computational efficiency.
We pose three key research questions to validate this hypothesis.
\textbf{RQ1:} \textit{From the perspective of in-context learning, what kind of data is good instruction tuning data?} \textbf{RQ2:} \textit{Are samples that are difficult for a model necessarily strong demonstrations or tuning examples?} \textbf{RQ3:} \textit{Do examples that yield high in-context influence also lead to superior instruction-following performance when used for fine-tuning?}

To answer these questions, we introduce a weighted in-context influence framework for data selection. For each candidate example we build a probe set in three stages that ensures semantic relatedness, diversity, and challenge. We then measure single-probe influence as the absolute reduction in instruction-following difficulty when the candidate is used as a one-shot demonstration. These influences are aggregated with a normalized cosine-distance weight to emphasize transferable gains. Finally, we rank by the aggregated score and select examples greedily under a diversity constraint on cosine similarity.
By quantifying each example's  influence on peers, we select a subset of examples that maximally improves overall instruction-following under a constrained data budget.

Our contributions are threefold. First, we introduce a weighted in-context influence framework that evaluates samples through dynamic probe sets and difficulty weighting rather than fixed global evaluation. 
Second, we demonstrate that local peer influence outperforms global assessment and that sample difficulty negatively correlates with teaching effectiveness. 
Third, extensive experiments show our method consistently outperforms existing baselines across multiple  benchmarks while achieving substantial computational efficiency.

\section{Preliminary}

In this section, we formalize our problem setup of data selection for instruction tuning, 
and introduce the concepts of perplexity and its ratio-based form, instruction–following difficulty (IFD), 
as intrinsic measures of sample difficulty.

\paragraph{Data Selection of Instruction Tuning} Let $D = \{(x_i, y_i)\}_{i=1}^n$ be an instruction–response corpus, where $x_i = T(\textit{Instruction},[\textit{Input}])$ denotes the full prompt produced by applying a prompt template $T$ to the instruction (and optional input) and the reference response $y_i$. The average negative log-likelihood $\mathcal{L}$ of an LLM $f_{\boldsymbol\theta}$ on $D$ is defined as:
\begin{equation}
\mathcal{L}(\boldsymbol\theta; D) = -\frac{1}{n} \sum_{i=1}^n \log p_{\boldsymbol\theta}(y_i \mid x_i).
\end{equation}

A lower $\mathcal{L}$ indicates stronger instruction-following ability.
Given the data budget $k$ , our goal is to select a subset $Q \subseteq D$ such that $|Q| = k$, after instruction tuning on $Q$, the resulting model achieves the strongest instruction-following ability:
\begin{equation}
Q^\star = \underset{Q \subseteq D,\, |Q| = k}{\arg\min} \; \mathcal{L}(\boldsymbol\theta_Q; D_{\text{test}}).
\end{equation}

\paragraph{Perplexity and Instruction–Following Difficulty (IFD).} 
For a sample of response length \(N\), 
perplexity is defined as:

{\small
\begin{equation}
\text{PPL}(y_i|x_i) = \exp\left(
-\frac{1}{N}\sum_{j=1}^{N}\log p(y_{i,j}|x_i, y_{i,1}, ..., y_{i,j-1})\right) 
\end{equation}
}

The $\mathcal{L}$ refers to instruction-following ability and its exponential form, perplexity, an ideal method for measuring the difficulty of generating the response.
Following~\citep{li2024quantity}, we measure
\textit{instruction–following difficulty} (IFD) as the benefit the
instruction provides over unconditional generation:
where a larger IFD indicates that the model gains less from the instruction.
\begin{equation}
\label{eq:ifd}
\text{IFD}(y\mid x)=
\frac{\text{PPL}(y\mid x)}{\text{PPL}(y)},
\end{equation}

\section{Method}
\begin{figure*}[htbp]
  \centering
  \includegraphics[width=\textwidth]{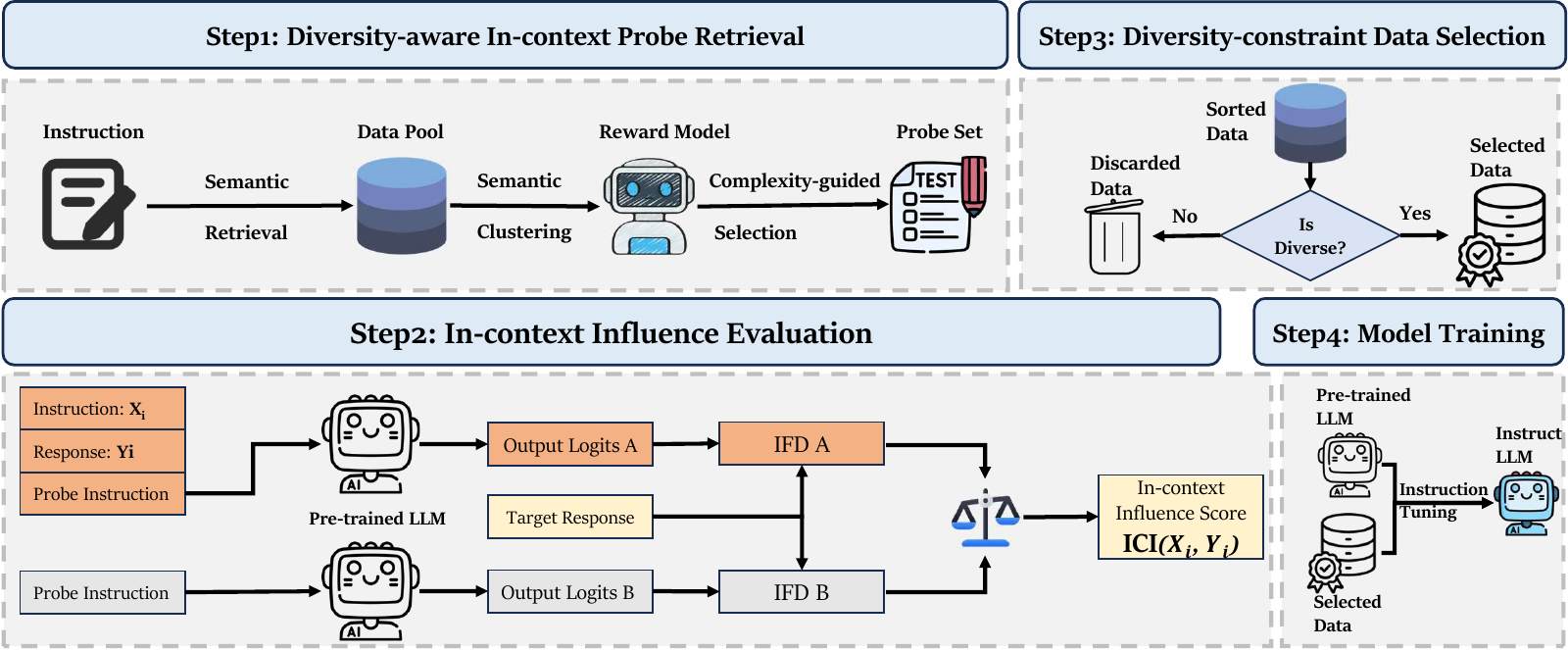}
  \caption{Overview of our framework.}
  \label{fig:framework}
\end{figure*}

In this section,  we present the proposed methodology in Figure~\ref{fig:framework}.
The key idea follows our hypothesis: \textit{In-context demonstrations that substantially reduce the instruction-following difficulty of challenging tasks are likely to be high-quality samples for instruction tuning.}
Our method consists of four major stages: 1) diversity-aware in-context probe retrieval, 2) in-context influence evaluation, 3) diversity-constraint data selection and 4) model training. 

\subsection{Diversity-aware In-context Probe Retrieval}

To assess how much a candidate demonstration helps related examples in in-context learning, we first construct a high-quality probe set.  
A naive retrieval of random or nearest-neighbor samples often fails to capture a demonstration’s true utility: unrelated probes introduce noise, redundant probes provide limited additional signal, trivially simple probes obscure the model’s genuine capability to follow complex instructions.
To overcome these limitations, we design a three-stage retrieval process consisting of Semantic Retrieval, Semantic Clustering, and Complexity-guided Selection, which ensures that the probe set is semantically relevant, diverse in coverage, and sufficiently challenging for evaluation.

\paragraph{Semantic Retrieval.}
In-context influence can only be measured meaningfully when the probes share a coherent topical space with the instruction; otherwise, their responses may reflect task mismatch rather than demonstrational assistance.  
For each instruction \(x_i\), we retrieve its \(N\) nearest neighbors in the embedding space under the Euclidean distance:
\begin{equation}
\mathcal{N}_i^{N}
=\operatorname*{arg\,topN}_{j\neq i}
\bigl(-\ell_2(f(x_i),f(x_j))\bigr).
\label{eq:nn_retrieval}
\end{equation}
where \(f(\cdot)\) denotes the sentence encoder.  

\paragraph{Semantic Clustering.}
While semantic retrieval enforces relevance, it tends to produce highly redundant neighbors concentrated in a narrow region of the embedding space.  
Such redundancy weakens the generality of influence estimation and biases evaluation toward repetitive behaviors.  
To mitigate this, we apply \(K\)-means clustering to partition \(\mathcal{N}_i^{N}\) into \(K\) semantically distinct groups.
Each cluster corresponds to a local semantic mode, thereby ensuring that the probe set explores diverse contexts related to \(x_i\).

\paragraph{Complexity-guided Selection.}
Finally, the probe set must include examples that are sufficiently challenging to reveal genuine improvements in instruction following.  
As observed by \citet{liu2022makes}, trivially simple probes provide little information about a demonstration’s true capacity, whereas complex probes expose whether it can genuinely lower instruction-following difficulty.  
To operationalize this insight, we employ a reward model \(R(\cdot)\) ~\footnote{https://huggingface.co/hkust-nlp/deita-complexity-scorer} that estimates instruction complexity.  
Within each cluster, we rank candidates by \(R(x_j)\) and select the highest-scoring instance:
\begin{equation}
\mathcal{B}_i=
\bigcup_{c=1}^{K}
\operatorname*{arg\,max}_{x_j\in\text{cluster}_c} R(x_j).
\label{eq:probe_set}
\end{equation}

The resulting probe set \(\mathcal{B}_i\) integrates semantic relevance, diversity, and challenge, forming a robust basis for subsequent in-context influence evaluation.

\subsection{In-context Influence Evaluation}
\label{sec:ici}

While instruction–following difficulty (IFD) serves as an intrinsic indicator of how hard an instruction is for a model to complete, 
it does not capture how one example can \emph{influence} the model’s behavior on other tasks during in-context learning.  
To address this limitation, we propose a new metric, the \textbf{In-context Influence (ICI)}, 
which quantifies how much a candidate demonstration improves the model’s instruction-following performance on its probes.
Given a candidate demonstration $a_i=(x_i,y_i)$ and its probe set $\mathcal{B}_i=\{(x_b,y_b)\}$ obtained from Step~1,  
we define the in-context influence on a single probe $b\in\mathcal{B}_i$ as the absolute reduction in instruction–following difficulty (IFD) when $a_i$ is provided as an in-context example:

{\small
\begin{equation}
\mathrm{ICI}_{i\rightarrow b}
=\text{IFD}(y_b\mid x_b)
-\text{IFD}(y_b\mid a_i,x_b).
\label{eq:ici_single}
\end{equation}}

A positive $\mathrm{ICI}_{i\rightarrow b}$ indicates that the demonstration $a_i$ reduces the difficulty of probe $b$, while a negative value reflects detrimental influence.
To explicitly encourage generalization beyond trivial semantic similarity, we weight each single-probe influence by a
normalized cosine distance that is \emph{monotonically increasing} with semantic distance.
We then define the overall weighted in-context influence \textit{(wICI)} as:

{\small
\begin{equation}
\mathrm{wICI}(a_i)
=\sum_{b\in\mathcal{B}_i}
\frac{1-\cos \! \bigl(f(x_i),f(x_b)\bigr)}{2\,|\mathcal{B}_i|}\;
\mathrm{ICI}_{i\rightarrow b}.
\label{eq:wici}
\end{equation}}

\subsection{Diversity-constraint Data Selection and Model Training}

To ensure that the selected data not only have high informativeness but also cover diverse semantic patterns of target samples, we introduce a diversity-constraint mechanism during data selection. 
Once every data candidate has a $\mathrm{wICI}$ score, we sort candidates in descending order and greedily build the coreset by adding a candidate $a_i=(x_i,y_i)$ only if it is not overly similar to any already selected item; otherwise it is skipped. Let $f(\cdot)$ be the sentence encoder and $\tau$ a cosine-similarity threshold; the admission test is

{\small
\begin{equation}
\max_{a_j\in\mathcal{S}} \cos\!\bigl(f(x_i),f(x_j)\bigr) < \tau,
\end{equation}}

and we continue until $|\mathcal{S}|=k$.
The resulting coreset is then used, without any additional weighting, to fine-tune the pre-trained LLM to improve the instruction following ability.

\section{Experimental Settings}

\subsection{Training Datasets}
We use two open-source instruction-tuning datasets: Alpaca-GPT4~\citep{peng2023instructiontuninggpt4}, which is synthesized through the Self-Instruct~\citep{wang2023self-instruct} paradigm and provides diverse, higher quality instruction–response pairs, and WizardLM~\citep{xu2023wizardlmempoweringlargelanguage}, which applies instruction evolution to expand basic prompts into harder multi step tasks to strengthen compositional reasoning. 
We adopt Wizard-70K and follow \citet{li2024quantity} to filter low quality samples. 
For fairness, we cap the training data at 10\% for our method and all baselines.

\subsection{Baselines}

\paragraph{Superfiltering (IFD)~\citep{li2024superfiltering}} observes a strong consistency in the Instruction-Following Difficulty (IFD) between small and large language models when applied to the same data. Based on this insight, it adopts a weak-to-strong paradigm: a smaller model is used to compute IFD scores and select data, which is then used to fine-tune a larger model. 
\paragraph{DEITA~\citep{liu2024what}} identifies three key factors in data selection: instruction complexity, instruction-response quality, and data diversity. It employs two reward models distilled from proprietary LLMs to assess instruction complexity and instruction-response quality, respectively. Additionally, it incorporates a diversity constraint during the data selection process to ensure a more varied dataset.

\paragraph{NUGGETS~\citep{li2024oneshot}} evaluates each instruction example's potential to serve as an effective one-shot demonstration by measuring its impact on model performance across a diverse anchor set of tasks. Specifically, NUGGETS computes a ``golden score'' for each instruction by comparing the model's zero-shot performance against its one-shot performance when using that instruction as context.

\paragraph{SelectIT~\citep{liu2024selectit}} exploits intrinsic uncertainty in LLMs to select high-quality instruction tuning data through three levels of self-reflection: token-level confidence analysis, sentence-level prompt variance reduction, and model-level collaborative assessment.

\subsection{Evaluation Setup}
\subsubsection{Pair-wise Comparison}
To assess the effectiveness of different data selection strategies, we perform pair-wise comparisons between models fine-tuned on selected 10\% subsets of data and a reference model fine-tuned on the full data. 
We use five publicly available test sets for evaluation: WizardLM~\citep{xu2023wizardlmempoweringlargelanguage}, Self-Instruct~\citep{wang2023self-instruct}, Vicuna~\citep{vicuna2023}, Koala~\citep{vu2023koala}, and LIMA~\citep{lima}. For each instruction in these datasets, both the full-data model and the subset-trained model are prompted to generate responses. These outputs are then evaluated by a strong LLM judge~\footnote{We use GPT-4.1-mini as the judge. For the prompt used for evaluation, please refer to the Appendix.}, which compares the two responses and assigns a preference score.

To reduce position bias, each response pair is presented to the judge in both possible orders. 
A model is considered the winner for a given instruction if its response is preferred or not worse in both orders. 
Based on this, we compute a \textit{winning score} for each method to quantify its relative performance against the full-data baseline. The score is defined as:

{\small
\begin{equation}
\text{winning\_score}(\mathcal{D}_t) = \frac{\text{num(wins)} - \text{num(loses)}}{\text{num}(\mathcal{D}_t)} + 1
\end{equation}
}

\subsubsection{Benchmark Evaluation}

In addition to pair-wise comparison, we evaluate models on a set of widely used benchmark datasets including ARC-Challenge~\citep{allenai:arc}, HELLASWAG (HS)~\citep{zellers2019hellaswag}, MMLU~\citep{hendrycks2021measuring}, BBH~\citep{bbh}, GSM8K~\citep{cobbe2021gsm8k}, MT-bench~\citep{zheng2023judging}
and 2Wikimqa, HotpotQA, Samsum from LongBench~\citep{bai2025longbench}.
Additionally, we report results on the AlpacaEval 2.0~\citep{dubois2024length} leaderboard using both Win Rate (WR) and Length-Controlled Win Rate (LC) metrics.

\begin{table}[!hbpt]
  \centering
  \resizebox{\columnwidth}{!}{%

    \begin{tabular}{@{}llcc@{}}
      \toprule
      \textbf{Dataset} & \textbf{Strategy}
        & \textbf{Llama3.1-8B}
        & \textbf{Mistral-7B} \\
      \midrule
      \multirow{6}{*}{Alpaca-GPT4}
        & Full           & 1.000 & 1.000 \\
        & IFD            & 1.198 & 1.248 \\
        & DEITA          & 1.076 & 1.099 \\
        & NUGGETS        & 1.133 & 1.201 \\
        & SelectIT       & 1.146 & 1.227 \\
        & Ours           & \textbf{ 1.215}     & \textbf{1.261 }    \\
      \midrule
      \multirow{6}{*}{WizardLM}
        & Full           & 1.000 & 1.000 \\
        & IFD            & \textbf{1.186} & 1.294 \\
        & DEITA          & 1.114 & 1.140 \\
        & NUGGETS        & 1.133 & 1.249 \\
        & SelectIT       & 1.176 & 1.281 \\
        & Ours           & 1.169 & \textbf{1.308} \\
      \bottomrule
    \end{tabular}%
   }
  \caption{Pairwise evaluation performance using Llama3.1-8b and Mistral-7B-v0.3 respectively.}
  \label{tab:winning}
\end{table}

\begin{table*}[!ht]
  \centering
  \small
  \setlength{\tabcolsep}{1.3mm}
  \begin{tabular}{l c c c c c c c c c c c}
    \toprule 
   & ARC-C & HS & MMLU & BBH & GSM8k & 2Wikimqa & HotpotQA
   & Samsum & MT-Bench & \multicolumn{2}{c}{AE 2.0} \\
    & Acc & Acc & Acc & EM & EM & F1 &  F1
    & ROUGE-L & Score & WR & LC \\
    \midrule
    \multicolumn{12}{c}{\cellcolor{gray!30}Llama3.1-8B w/ Alpaca-GPT4} \\
    {\color[HTML]{656565}Full}   
      & {\color[HTML]{656565}52.99}
      & {\color[HTML]{656565}79.78}
      & {\color[HTML]{656565}61.81}
      & {\color[HTML]{656565}40.90}
      & {\color[HTML]{656565}47.46}
      & {\color[HTML]{656565}39.44}
      & {\color[HTML]{656565}41.73}
      & {\color[HTML]{656565}36.03}
      & {\color[HTML]{656565}4.30}
      & {\color[HTML]{656565}5.65} & {\color[HTML]{656565}13.19} \\
    IFD     & 53.50 & 80.95 & 63.37 & 35.52 & 53.42 & \textbf{43.52} & 44.28 & 36.21 & 4.84 &  \textbf{7.58} & \textbf{14.85} \\
    DEITA   & 58.21 & 80.46 & 62.77 & \textbf{40.92} & 52.05 & 42.99 & 42.12 & 35.11 & 4.68 &  5.27 & 12.14 \\
    NUGGETS & 57.42 & 80.76 & 62.89 & 39.91 & 52.77 & 43.35 & 42.92 & 34.81 & 4.55 &  5.85 & 13.03 \\
    SelectIT& 58.16 & 81.24 & 62.56 & 39.93 & 53.90 & 42.03&44.48&35.10& 4.73 &  6.22 & 13.90 \\

    Ours &\textbf{58.98}&\textbf{81.52}&\textbf{63.45}&39.33&\textbf{55.17}&43.47 & \textbf{44.74} & \textbf{36.60}&\textbf{4.88}&7.50&14.42 \\
    \midrule
    \multicolumn{12}{c}{\cellcolor{gray!30}Llama3.1-8B w/ Wizard} \\
    {\color[HTML]{656565}Full}    
      & {\color[HTML]{656565}54.61}
      & {\color[HTML]{656565}78.36}
      & {\color[HTML]{656565}61.32}
      & {\color[HTML]{656565}42.20}
      & {\color[HTML]{656565}55.42}
      & {\color[HTML]{656565}45.71}
      & {\color[HTML]{656565}56.10}
      & {\color[HTML]{656565}30.09}
      & {\color[HTML]{656565}4.75}
      & {\color[HTML]{656565}6.02} & {\color[HTML]{656565}14.75}   \\
    IFD     & 56.40 & 80.42 & 63.65 & 40.67 & 50.64 & 47.06 & 55.03 & 30.05 & \textbf{5.41} & \textbf{6.34} & 12.82 \\
    DEITA   & 57.29 & 79.19 & 63.52 & 38.70 & 51.83 & 46.37 & 56.02 & 31.63 & 5.23 &  5.90 & 11.42 \\
    NUGGETS & 56.50 & 79.08 & 64.04 & 39.71 & 50.22 & 43.35 & 53.43 & \textbf{32.68} & 4.91 &  5.74 & 11.71 \\
    SelectIT& 57.17 & 80.47 & 64.11 & 40.54 & 51.03 & 45.99 & 56.19 & 30.55 & 5.02 &  5.57 & 11.43 \\

    Ours &\textbf{57.79}&\textbf{81.02}&\textbf{64.90}&\textbf{41.21} &\textbf{52.84}&\textbf{47.06}&\textbf{57.14}&32.06&5.28&6.11&\textbf{13.13} \\
    \midrule
    \multicolumn{12}{c}{\cellcolor{gray!30}Mistral-7B-v0.3 w/ Alpaca-GPT4} \\
    {\color[HTML]{656565}Full}    
      & {\color[HTML]{656565}44.03}
      & {\color[HTML]{656565}73.01}
      & {\color[HTML]{656565}51.40}
      & {\color[HTML]{656565}31.18}
      & {\color[HTML]{656565}18.73}
      & {\color[HTML]{656565}37.65}
      & {\color[HTML]{656565}37.00}
      & {\color[HTML]{656565}21.91}
      & {\color[HTML]{656565}3.80}
      & {\color[HTML]{656565}5.65} & {\color[HTML]{656565}13.19}     \\
    IFD     & 48.81 & 78.42 & \textbf{55.50} & 33.68 & 26.80 & 42.11 & 34.54 & 33.16 & 3.98 &  \textbf{6.50} & \textbf{11.92} \\
    DEITA   & 48.33 & 80.52 & 54.14 & 33.70 & 27.43 & \textbf{42.33} & 34.54 & 33.91 & 4.14 &  5.92 & 11.02 \\
    NUGGETS & 48.25 & 80.33 & 53.29 & 33.52 & 27.99 & 39.89 & 31.05 & 31.41 & 4.22&  5.10 & 10.84 \\
    SelectIT& 49.01 & 80.21 & 54.04 & 33.44 & 28.26 & 40.54 & 32.72 & 32.49 & \textbf{4.32} &  6.12 & 11.35 \\    

    Ours &\textbf{49.43}&\textbf{81.14}&54.73&\textbf{34.24}&\textbf{28.53}&41.57&\textbf{34.92}&\textbf{34.50}&4.18&6.26&11.35 \\
    \midrule
    \multicolumn{12}{c}{\cellcolor{gray!30}Mistral-7B-v0.3 w/ Wizard} \\
    {\color[HTML]{656565}Full}    
      & {\color[HTML]{656565}46.25}
      & {\color[HTML]{656565}73.57}
      & {\color[HTML]{656565}51.15}
      & {\color[HTML]{656565}31.84}
      & {\color[HTML]{656565}32.37}
      & {\color[HTML]{656565}44.76}
      & {\color[HTML]{656565}48.85}
      & {\color[HTML]{656565}21.16}
      & {\color[HTML]{656565}3.97}
      & {\color[HTML]{656565}4.60} & {\color[HTML]{656565}10.77}     \\
    IFD     & 50.08 & 78.39 & 55.99 & 36.55 & 28.89 & \textbf{42.88} & \textbf{47.40} & 29.08 & 4.15 & 5.84 &  9.91 \\
    DEITA   & 47.76 & 76.25 & 54.81 & \textbf{37.18} & 28.85 & 41.91 & 46.80 & 28.66 & 4.35 &  5.09 & 11.32 \\
    NUGGETS & 50.23 & 77.36 & 55.69 & 36.37 & 27.64 & 40.43 & 44.77 & 28.02 & 4.32 &  4.88 & 10.51 \\
    SelectIT& 51.02 & 78.43 & 55.10 & 37.09 & 28.01 & 41.27 & 47.33 & 28.78 & 4.10 &  5.25 & 11.18 \\      

    Ours &\textbf{51.27}&\textbf{78.51} &\textbf{56.31} &36.99&\textbf{29.44}&42.47&46.53&\textbf{29.22}&\textbf{4.40}&\textbf{5.91}&\textbf{11.36} \\
    \bottomrule
  \end{tabular}
  
  \caption{Benchmark zero-shot evaluation of data‐selection methods using 10\% of the training data. {\color[HTML]{656565}``Full''} denotes performance on the entire dataset for reference. WR and LC stand for Winning Rate and Length-Controlled winning rate, respectively.}
  \label{tab:main}
\end{table*}

\section{Main Experimental Results (RQ1)}

\subsection{Pairwise Evaluation Results}

The pairwise evaluation in Table \ref{tab:winning} shows that our data selection yields consistent and substantial gains, achieving competitive or superior performance compared to existing baselines across models and datasets.
Relative to ``Full'' our method improves the winning score by  +21.5\% and +26.1\% on Alpaca-GPT4 when fine‐tuning Llama3.1-8B and Mistral, respectively.
On WizardLM, our method improves the winning score by +16.9\% and +30.8\% with Llama3.1-8B and Mistral, respectively. 
Overall, these results indicate that our data selection is highly effective across different models and datasets.

\subsection{Benchmark Results}

Table~\ref{tab:main} presents the zero-shot benchmark performance of models trained on the full dataset versus those fine-tuned on just 10\% of the data using baselines and our method.

First, the full-data models underperform the 10\% data-selection methods across nearly every evaluation, including both zero-shot benchmarks and the Alpaca-Eval~2.0 leaderboard, indicating substantial redundancy and noise in the original training set. 
Second, different data selection baselines exhibit task-specific strengths: on Alpaca-Eval~2.0, IFD outperforms DEITA, whereas on BBH, DEITA surpasses IFD. This indicates that different selection criteria steer model improvements along different capability dimensions rather than in a single unified direction.
Third, our budget-constrained selection matches or surpasses strong baselines and full-data models across zero-shot benchmarks and Alpaca-Eval 2.0. 
Although some baselines peak on individual metrics, our selection remains top two in nearly all settings, showing robust and broad effectiveness under a strict budget.
The final finding answers our \textbf{RQ1}: \emph{In-context demonstrations that substantially reduce the instruction-following difficulty of diverse challenging tasks are high-quality samples for instruction tuning.}

\section{In-Depth Analysis}

In this section we present a series of experiments and analyses to answer the rest two of our three research questions: \textbf{RQ2}: \emph{Are difficult data necessarily good demonstrations of in-context learning?} and \textbf{RQ3}: \emph{Are good in-context learning demonstrations good examples of instruction tuning? }

\subsection{Data Consistency Analysis (RQ2)}

To test whether samples that are difficult for the model are also effective in context demonstrations, we ranked every dataset twice, first by the instruction following difficulty (IFD) and then by unweighted in-context influence (ICI), and compared the two rankings.
We measured the overlap between the lists at the top 10\%, 30\%, and 50\% cut-offs and computed the Spearman correlation over the full lists. 
As shown in Table~\ref{tab:relation}, top-10\% overlap is limited at 10.1\% on Alpaca-GPT4 and 14.4\% on WizardLM, rising to about 38–39\% at the top 30\% and 59–65\% at the top 50\%.
Spearman correlations are positive but moderate at 0.39 and 0.26. 
These results indicate a meaningful yet incomplete alignment between instruction–following difficulty and in-context influence; thus, difficult samples are not reliably strong in-context demonstrations, answering \textbf{RQ2} in the negative.

\begin{figure*}[htbp]
  \centering
  \begin{subfigure}[b]{0.325\textwidth}
    \centering
    \includegraphics[width=\textwidth]{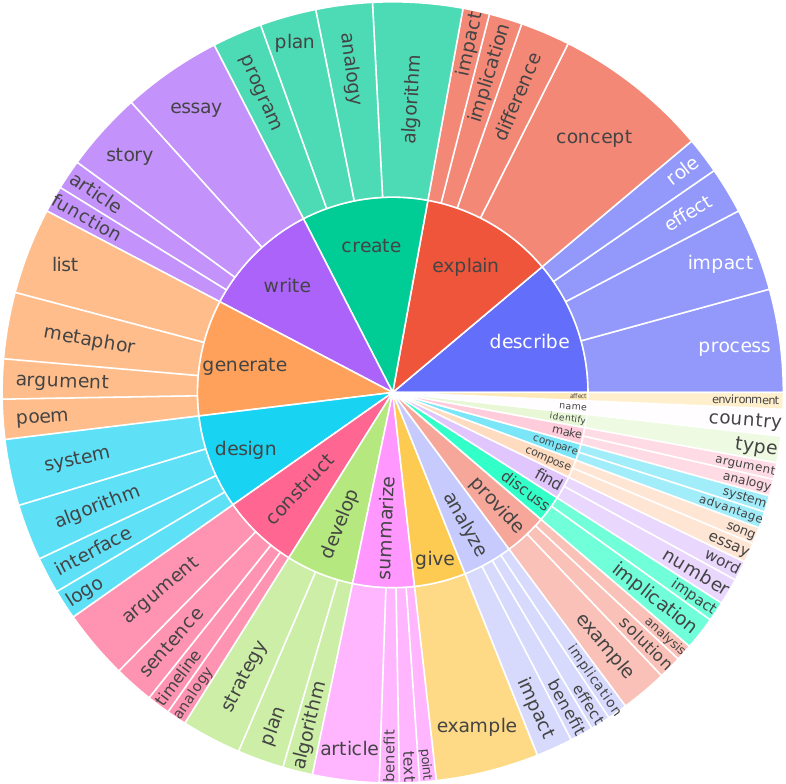}
    \caption{High Instruction Following Difficulty.}
    \label{fig:sub1}
  \end{subfigure}%
  \hspace{0.01\textwidth}%
  \begin{subfigure}[b]{0.325\textwidth}
    \centering
    \includegraphics[width=\textwidth]{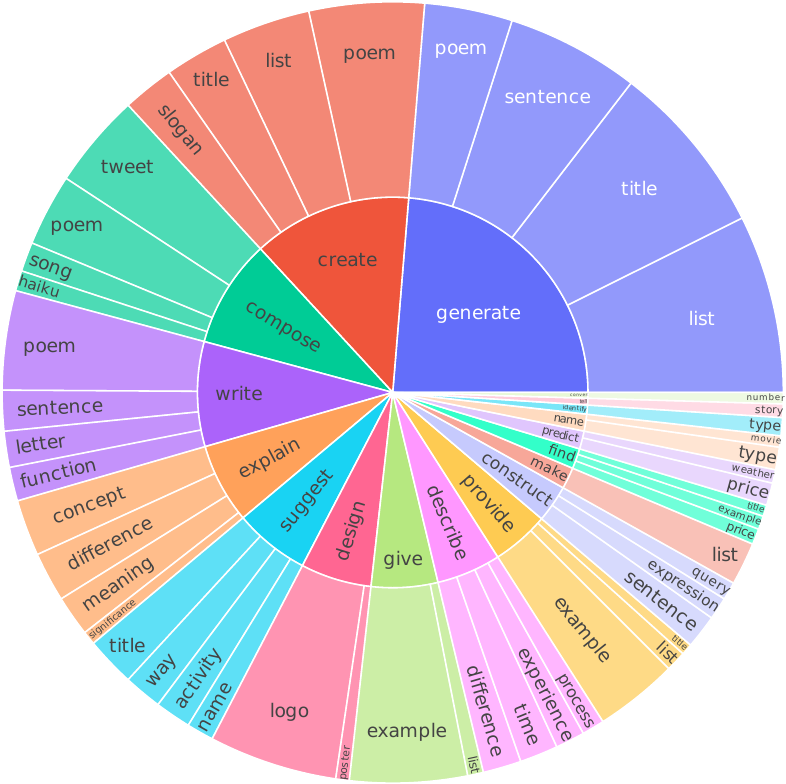}
    \caption{High In-context Influence.}
    \label{fig:sub2}
  \end{subfigure}%
  \hspace{0.01\textwidth}%
  \begin{subfigure}[b]{0.325\textwidth}
    \centering
    \includegraphics[width=\textwidth]{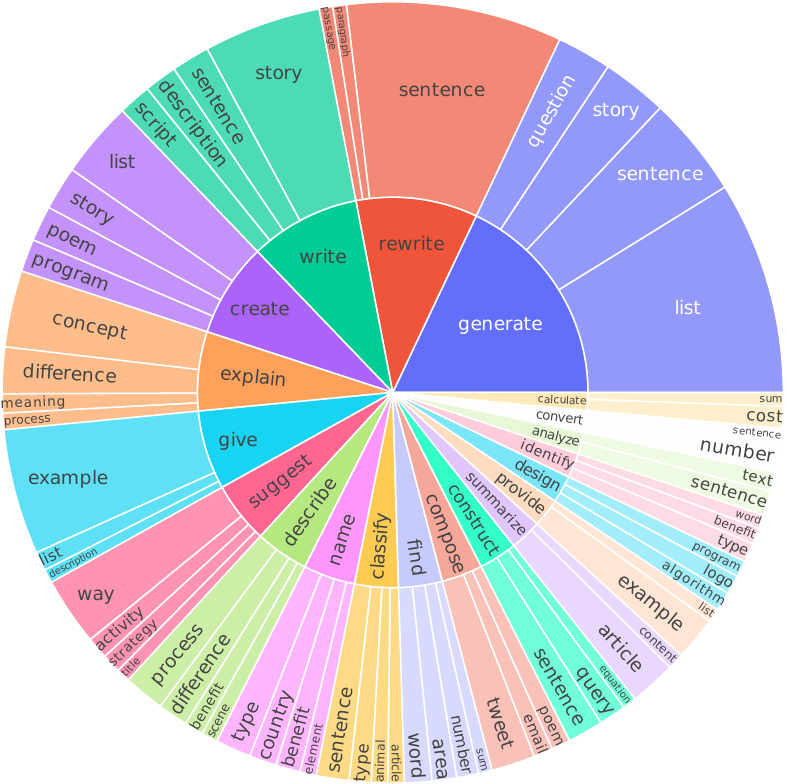}
    \caption{Random.}
    \label{fig:sub3}
  \end{subfigure}
  \caption{Visualization of the verb–noun structures in instructions selected by the three data-selection strategies on Alpaca-GPT4. The inner ring shows the predominant verbs, the outer ring displays the nouns that co-occur directly with those verbs.}
  \label{fig:cake}
\end{figure*}

Further, we adopt the approach of ~\citep{lou2024muffin} by using the Berkeley Neural Parser to extract root verbs and their direct objects from our extracted instruction subset. 
We then visualize the 20 most frequent root verbs alongside their four most common direct-object nouns under three conditions: (1) High Instruction Following Difficulty, (2) High In-context Influence, and (3) Random.
As shown in Figure~\ref{fig:cake}, the distributions differ substantially across these criteria, reinforcing our conclusion that sample difficulty and in-context learning utility capture distinct properties.

\begin{table}[!tbh]
\centering

\resizebox{0.98\columnwidth}{!}{%
\begin{tabular}{l|ccc|c}
\toprule
\multirow{2}{*}{\textbf{Dataset}} 
  & \multicolumn{3}{c|}{\textbf{Overlap Ratios} } 
  & \multirow{2}{*}{\textbf{Rank Corr.}} \\ 
  & 10\% & 30\% & 50\% &  \\ 

\midrule
Alpaca-GPT4 
  & 0.1006 & 0.3874 & 0.6476 & 0.3947 \\ 

\midrule
WizardLM
  & 0.1442& 0.3650 & 0.5942 & 0.2568 \\ 

\bottomrule
\end{tabular}%
}

\caption{Relation of high IFD  data and high ICI data.}
\label{tab:relation}
\end{table}

\subsection{Ablation Study (RQ3) }

We ablate two diversity components to quantify their contributions and answer RQ3. The variant \textit{w/o DA} removes Semantic Clustering in Step~1 (no probe-side diversity during influence estimation), and \textit{w/o DS} keeps influence scoring but drops the cosine-similarity constraint in selection (no demonstration-side diversity). As shown in Table~\ref{tab:ablation}, both variants underperform our full method across datasets and models, yet both still achieve winning scores greater than 1.0, which indicates they outperform training on the full data. 
Therefore, for \textbf{RQ3}: \emph{Are good in-context demonstrations good examples of instruction tuning?} the answer is \textbf{yes}, even without diversity modules they provide gains over the full-data baseline, however, they are most effective when coupled with diversity on both the probe side and the demonstration side.

\begin{table}[thp]
  \centering
  \resizebox{\columnwidth}{!}{%
    \begin{tabular}{@{}ll|cc@{}}
      \toprule
      \textbf{Dataset} & \textbf{Strategy}
        & \textbf{Llama3.1-8B}
        & \textbf{Mistral-7B} \\
      \midrule

      \multirow{3}{*}{Alpaca-GPT4}
        & w/o DA  & 1.140 &1.181  \\
        & w/o DS   & 1.155 &1.198  \\
        & Ours     &1.215  &1.261  \\
      \midrule
      \multirow{3}{*}{WizardLM}
        & w/o DA  &1.132  &1.204  \\
        & w/o DS   & 1.154 &1.239  \\
        & Ours     & 1.169 &1.308  \\
      \bottomrule
    \end{tabular}%
   }
  \caption{Ablation study on two major sample-selection components. \textit{w/o DA} removes Semantic Clustering in Step~1, \textit{w/o DS} drops the cosine-similarity constraint in selection.}
  \label{tab:ablation}
\end{table}

\subsection{Performance with Varied Budgets}

\begin{figure}
    \centering
    \includegraphics[width=\linewidth]{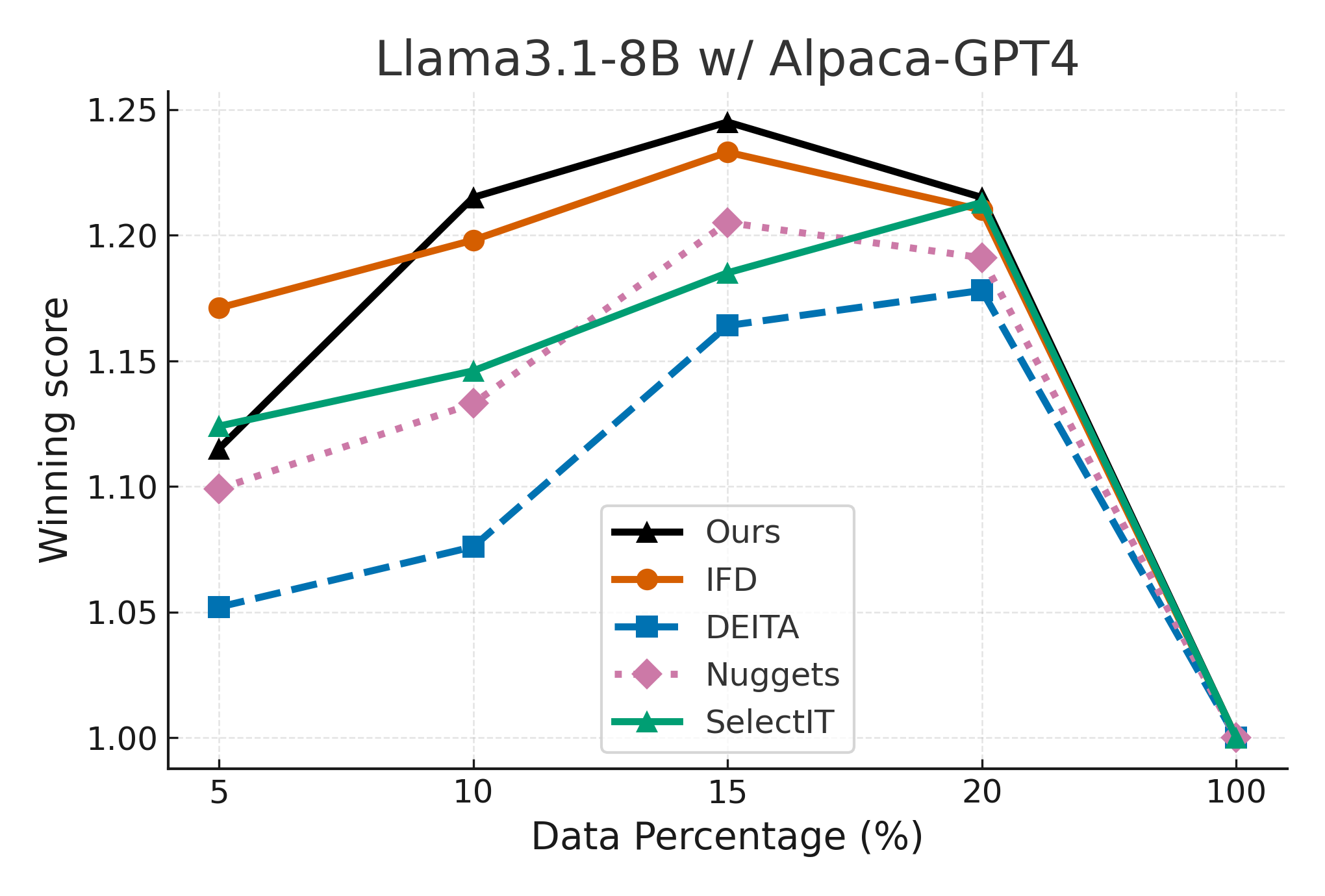}
    \caption{Winning-score curves of \textsc{Llama3.1-8B} trained on Alpaca-GPT4.}
    \label{fig:line}
\end{figure}

We next study how budget size influences each data selection method. 
We progressively raise the budget from five to twenty percent of the training pool and record the winning score of \textsc{Llama3.1-8B}.
As shown in Figure~\ref{fig:line}, our weighted-ICI method shows a steep performance climb between five and ten percent, reaches its apex at fifteen percent, and then drifts slightly downward as more data are added.  
Most baseline methods, including DEITA, NUGGETS, and SelectIT, follow similar curves but consistently lag behind our approach at every budget level, with IFD showing the closest performance to our method.
This behavior reveals two distinct phases of data addition. In the low-budget regime, carefully selected demonstrations introduce novel and complementary information, so increasing the data volume leads to better models. 
However, beyond a certain saturation point, additional examples tend to overlap with existing information or introduce noise, resulting in diminishing or even negative returns. 
Because our strategy prioritizes demonstrations based on their ability to aid the most challenging probes, it reaches the saturation threshold with far fewer samples, thereby achieving superior sample efficiency.

\subsection{Domain Generalization Analysis}

To examine whether our data selection method generalizes beyond the Alpaca-GPT4 and WizardLM settings, we additionally conduct experiments in a medical domain using the MedQuAD~\citep{ben2019question} dataset. Specifically, we select 30\% of the training data with our method, fine-tune LLMs on the resulting subset, and evaluate them on three medical benchmarks: MedMCQA~\citep{pal2022medmcqa}, MedQA~\citep{jin2021disease}, and MMLU-med. We compare our method against Random selection and Full-data fine-tuning. This setting provides a targeted test of whether the proposed wICI-based selection strategy remains effective when transferred to a domain that differs substantially from the general-purpose instruction-tuning corpora used in our main experiments.

Table~\ref{tab:medical_domain} reports the results. Overall, our method consistently outperforms Random selection across both model backbones on most evaluation metrics, showing that the selected subset remains substantially more effective than random 30\% sample even in the medical domain. Compared with Full-data fine-tuning, the results are more mixed: our method achieves the best performance on MedQA for Llama-3.1-8B and on both MedMCQA and MedQA for Mistral, while trailing the Full baseline on MMLU-med and on MedMCQA for Llama-3.1-8B. 
These results suggest that wICI generalizes beyond the domains considered in the main experiments and can still identify highly useful training subsets in a specialized domain. At the same time, the remaining gap on some settings indicates that full-data coverage may still be particularly valuable for certain types of domain-specific knowledge.

\begin{table}[t]
\centering
\small
\resizebox{\columnwidth}{!}{%
\begin{tabular}{lccc}
\toprule
& MedMCQA & MedQA & MMLU-med \\
\midrule
\multicolumn{4}{c}{\cellcolor{gray!30}Llama-3.1-8B} \\
\midrule
Full   & \textbf{40.53} & 44.22 & \textbf{71.33} \\
Random & 32.32 & 29.53 & 45.67 \\
Ours   & 39.63 & \textbf{46.03} & 65.33 \\
\midrule
\multicolumn{4}{c}{\cellcolor{gray!30}Mistral-7B-v0.3} \\
\midrule
Full   & 36.55 & 34.32 & \textbf{51.67} \\
Random & 33.11 & 37.00 & 36.00 \\
Ours   & \textbf{37.05} & \textbf{39.54} & 50.00 \\
\bottomrule
\end{tabular}%
}
\caption{Results on the medical domain. Our method and the Random baseline use 30\% of the training data, while Full uses the entire dataset. The best performance in each column is shown in bold.}
\label{tab:medical_domain}
\end{table}

\section{Related Work}

\paragraph{In-Context Learning for Instruction Tuning} 
In-context learning (ICL)~\citep{brown2020language} and instruction tuning (IT)~\citep{lou2024large} represent two fundamental paradigms for adapting large language models to downstream tasks. 
ICL enables models to perform tasks by conditioning on demonstration examples provided in the prompt, without updating model parameters.
In contrast, instruction tuning updates model parameters through supervised fine-tuning on instruction-response pairs to improve instruction-following capabilities.

Recent theoretical work has revealed deep connections between these paradigms. ~\citep{dai2023gpt} provided a groundbreaking theoretical explanation, demonstrating that Transformer attention has a dual form of gradient descent and that ICL can be understood as implicit fine-tuning where GPT produces meta-gradients according to demonstration examples. 
~\citep{von2023transformers} further demonstrated that transformers can implement learning algorithms implicitly, showing that this process corresponds to gradient-based optimization of principled objective functions. ~\citep{duan2024exploring} empirically validated these theoretical insights, finding that ICL changes an LLM's hidden states as if the demonstrations were used to instructionally tune the model, establishing that ``ICL is implicit IT.''
Building on these theoretical foundations, several works have explored practical applications.~\citep{li2024oneshot} made a pioneering contribution by proposing NUGGETS, which leverages the ICL-IT connection for practical data selection. NUGGETS evaluates instruction tuning candidates by measuring their effectiveness as one-shot demonstrations across a global anchor set, computing "golden scores" based on performance improvements.
\citep{mosbach2023shot} conducted a comprehensive comparison between few-shot fine-tuning and ICL, finding that both approaches achieve similar generalization performance when controlling for model size and training examples. 
In addition, related work has used ICL to synthesize instruction-tuning data.~\citep{han2025attributes,xu2023wizardlmempoweringlargelanguage}

Our work introduces In-Context Influence (ICI) to measure how effectively a candidate example reduces instruction-following difficulty for semantically related peers. Through local semantic relevance and difficulty weighting mechanisms, we achieve more precise and computationally efficient data selection than existing global evaluation approaches.

\paragraph{Instruction Data Selection}
Instruction-tuning corpora, whether in general-purpose settings or domain-specific scenarios~\citep{rao2024commonit,wei2025igniting,rao2026scoping}, often contain redundancy and noise, motivating the selection of compact, high-value subsets. Quality-based approaches score or mine examples, including AlpaGasus \citep{chen2024alpagasus}, DEITA \citep{liu2024what}, Superfiltering \citep{li2024superfiltering}, CherryLLM \citep{li2024quantity}, Instruction Mining \citep{cao2024instruction}, and the style-consistency method SCAR \citep{li2025scar}. 
Alignment-oriented filtering reduces hallucination, for example NOVA \citep{si2025nova}, and a call for rigor~\citep{moon2025rigor} highlights confounds from inconsistent training setups. 
Diversity-driven methods such as QDIT \citep{bukharin2024data} and DPP~\citep{wang2024diversity} enhance coverage and minimize redundancy through greedy or determinantal objectives. 
Information-theoretic and multi-signal approaches include MIG \citep{chen2025mig} for information gain, GRAPE \citep{zhang2025grape} for response fitness, ZIP \citep{yin2024entropylawstorydata} for entropy-based compression, and NICE \citep{wang2025nice} for task-level selection. A recent survey~\citep{liu2025rethinking} synthesizes these directions and proposes a unified framework.

Our work complements these lines by selecting examples that maximize in-context influence on related and challenging probes, measured as absolute reductions in instruction–following difficulty, and by enforcing diversity during probe construction and in the final coreset. 
By selecting examples according to their measured in-context influence, we align data selection with the actual inference-time benefit of demonstrations, producing compact yet transferable subsets while controlling redundancy through diversity.

\section{Conclusion}

We presented a data selection framework for instruction tuning based on a new influence metric. The metric, In-context Influence, measures the reduction in instruction-following difficulty caused by a candidate demonstration, and its generalization-weighted aggregate guides selection. 
Our pipeline builds probe sets that are related, diverse, and challenging, and applies a diversity constraint during selection to avoid redundancy. Under a 10\% budget, the method matches or surpasses strong baselines and full-data models across different benchmarks.
We also answered three research questions. \textbf{RQ1}: demonstrations that substantially reduce difficulty on diverse challenging probes are high-quality samples for instruction tuning. \textbf{RQ2}: difficult samples are not reliably strong in-context demonstrations. \textbf{RQ3}: good in-context demonstrations are effective instruction-tuning examples, and they work best when both probe diversity and demonstration diversity are enforced. 

\section*{Limitations}

Our study has two primary limitations. First, due to computational constraints, we did not evaluate larger backbones such as Llama3-70B~\citep{dubey2024llama}, nor did we test on substantially larger instruction corpora such as Tulu3~\citep{lambert2024tulu}. Second, we focused on supervised instruction tuning and did not evaluate other post-training methods such as DPO~\citep{rafailov2023direct}, PPO~\citep{schulman2017proximal} and their variants. 
We expect our selection framework to transfer to larger models and alternative post-training methods, which we leave to future work.

\section*{Acknowledgment}
The authors thank anonymous reviewers for their insightful feedback. 
The project was partially supported by the National Science Foundation (NSF) under awards CNS-2318210 and TI-2434589 (OpenAI API expenses).
We thank the computing resources provided by the iTiger GPU cluster~\cite{sharif2025ITIGER} supported by the NSF MRI program under the award CNS-2318210.

\bibliography{main}

\appendix

\newpage
\section{Implementation Details}

\subsection{Hyperparameters}
We use $k\!=\!32$ nearest neighbors for semantic retrieval, $K\!=\!5$ clusters for semantic clustering (k-means on $\ell_2$-normalized embeddings), and a cosine-similarity threshold $\tau\!=\!0.9$ to enforce diversity during demonstration selection.

All experiments in this paper adopt a unified configuration. We use Llama Factory~\citep{zheng2024llamafactory} to fully fine-tune Llama3.1-8B, and Mistral-7B-v0.3. 
Fine-tuning is performed with DeepSpeed ZeRO-3 for memory optimization and bf16 mixed precision, using input sequences truncated to 2,048 tokens. Each model is trained for three epochs with the AdamW optimizer, an initial learning rate of $1 \times 10^{-5}$, a cosine-annealing learning rate schedule with linear warmup at a rate of 0.1, and a total batch size of 64.

The prompt template of the reward model is shown in Figure~\ref{fig:complexity_prompt}.

\begin{figure*}[!t]
\small
\begin{verbatim}
System Prompt:
You are a helpful assistant. Please identify the complexity score
of the following user query.

User Prompt:
##Query:
{instruction}

##Complexity:
\end{verbatim}
\caption{Prompt template used to elicit a discrete complexity level (1--6) from the DEITA complexity scorer. 
At inference, we read the logits for the six numeral tokens and compute the expected score 
$\sum_{k=1}^{6} k \cdot \mathrm{softmax}(\ell_k)$, where $\ell_k$ is the logit of token $k$.}
\label{fig:complexity_prompt}
\end{figure*}

\subsection{Hardware and Software}

We conducted all experiments on a machine equipped with 8x 6000Ada GPUs, 2x EPYC Genoa 9334 CPUs, and 768GB of RAM. 
The system runs on Linux kernel 5.14.

For the Sentence Transformer library~\citep{reimers2019sentence}  used in our experiments, we utilize the \texttt{stsb-roberta-large} checkpoint as the embedding model. 
\section{Evaluation Details}
\subsection{Benchmark Evaluation}
For evaluation on standard NLP benchmarks, we use the lm-evaluation-harness~\citep{eval-harness}, following their default zero-shot settings.

\subsection{Pairwise Evaluation}
We employ GPT-4.1-mini as a judge for pairwise evaluation using the prompt template shown in Figure~\ref{fig:eval_prompt}.

\begin{figure*}[!t]
\small
\begin{verbatim}
System Prompt:
You are a helpful and precise assistant for checking the quality of the answer.

User Prompt:
[Question]
{question}

[The Start of Assistant 1's Answer]
{ans1}
[The End of Assistant 1's Answer]

[The Start of Assistant 2's Answer]
{ans2}
[The End of Assistant 2's Answer]

[System]
We would like to request your feedback on the performance of two AI 
assistants in response to the user question displayed above. Please 
rate the helpfulness, relevance, accuracy, level of details of their 
responses. Each assistant receives an overall score on a scale of 1 
to 10, where a higher score indicates better overall performance. 
Please first output a single line containing only two values indicating 
the scores for Assistant 1 and 2, respectively. The two scores are 
separated by a space. In the subsequent line, please provide a 
comprehensive explanation of your evaluation, avoiding any potential 
bias and ensuring that the order in which the responses were presented 
does not affect your judgment.
\end{verbatim}
\caption{Prompt template used for pairwise evaluation with GPT-4.1-mini as judge.}
\label{fig:eval_prompt}
\end{figure*}

\section{Efficiency Analysis}
\begin{table*}[!ht]
\centering
\resizebox{\textwidth}{!}{%
\begin{tabular}{@{}l|cccc@{}}
  \toprule
  \textbf{Method} & \textbf{Backward} & \textbf{Teacher Model} & \textbf{External Knowledge} & \textbf{Number of Forward Passes } \\
  \midrule
  IFD              & \xmark & \xmark & \xmark & 2/sample (with/without instruction) \\
  DEITA            & \cmark & \cmark & \xmark & 36,000 (reward model training) + 2/sample (model scoring)  \\
  NUGGETS          & \xmark & \xmark & \xmark & 2,000/sample (anchor set evaluation) \\
  SelectIT         & \xmark & \cmark & \xmark & 15/sample (multi-prompt, multi-LLM) \\
  RECOST           & \xmark & \xmark & \cmark & 2/sample (with/without external KB) \\
  \midrule
  \textbf{Ours}    & \xmark & \xmark & \xmark & 16/sample (5 probes × 3 + 1) \\
  \bottomrule
\end{tabular}
}
 \caption{Computational Efficiency Analysis of Data Selection Methods}
 \label{tab:efficiency}
\end{table*}
To better understand the practical implications of our approach, we analyze the computational efficiency of different data selection methods. Table~\ref{tab:efficiency} provides a comprehensive comparison across several dimensions: backward pass requirements,  dependency on teacher models, external knowledge requirements, and the number of forward passes needed per sample.

Our analysis reveals significant differences in computational overhead across methods. 
DEITA incurs the highest cost, requiring 36,000 forward passes to train reward models plus 2 forward passes per sample for scoring. 
This substantial training overhead makes DEITA particularly expensive for large-scale data selection scenarios. 
NUGGETS, while avoiding the training cost, requires 2,000 forward passes per sample to evaluate each candidate against its fixed anchor set, resulting in prohibitive computational costs when processing large datasets.
In contrast, uncertainty-based methods like SelectIT (15 forward passes per sample) and RECOST (2 forward passes per sample) offer more reasonable computational requirements. 
However, SelectIT's multi-prompt and multi-LLM evaluation strategy still introduces considerable overhead, while RECOST's dependency on external knowledge bases may limit its applicability in certain domains.

Our method strikes a balance between evaluation precision and computational efficiency. With 16 forward passes per sample (5 probes × 3 evaluations + 1 for the candidate itself), our approach requires significantly fewer computations than NUGGETS while maintaining the benefits of contextual evaluation. Unlike DEITA, our method requires no training phase, and unlike RECOST, it operates without external dependencies. The dynamic nature of our probe selection ensures that computational resources are focused on the most relevant semantic neighbors, maximizing the informativeness of each evaluation.
This efficiency analysis demonstrates that our weighted in-context influence framework achieves superior performance gains while maintaining practical computational requirements, making it suitable for real-world data selection scenarios at scale.

\begin{table*}[ht]
\centering
\begin{tabular}{lcccccc}
\toprule
Model / Dataset & Full & IFD & DEITA & NUGGETS & SelectIT & Ours \\
\midrule
Llama3.1-8B / Alpaca-GPT4  & \textbf{48.83} & 42.31 & 36.87 & 38.55 & 34.11 & 42.37 \\
Llama3.1-8B / WizardLM     & \textbf{53.01} & 49.60 & 49.05 & 50.53 & 52.11 & 51.19 \\
Mistral-7B / Alpaca-GPT4   & \textbf{41.74} & 35.19 & 33.70 & 32.80 & 32.77 & 38.35 \\
Mistral-7B / WizardLM      & \textbf{50.92} & 48.48 & 48.13 & 48.15 & 47.75 & 48.43 \\
\bottomrule
\end{tabular}
\caption{IFEval benchmark evaluation. We report the average of four metrics: Pr(S), Pr(L), Ins(S), and Ins(L). The best performance in each row is shown in bold.}
\label{tab:ifeval}

\end{table*}

\begin{figure*}[htpb]
  \centering
  \begin{subfigure}[b]{0.325\textwidth}
    \centering
    \includegraphics[width=\textwidth]{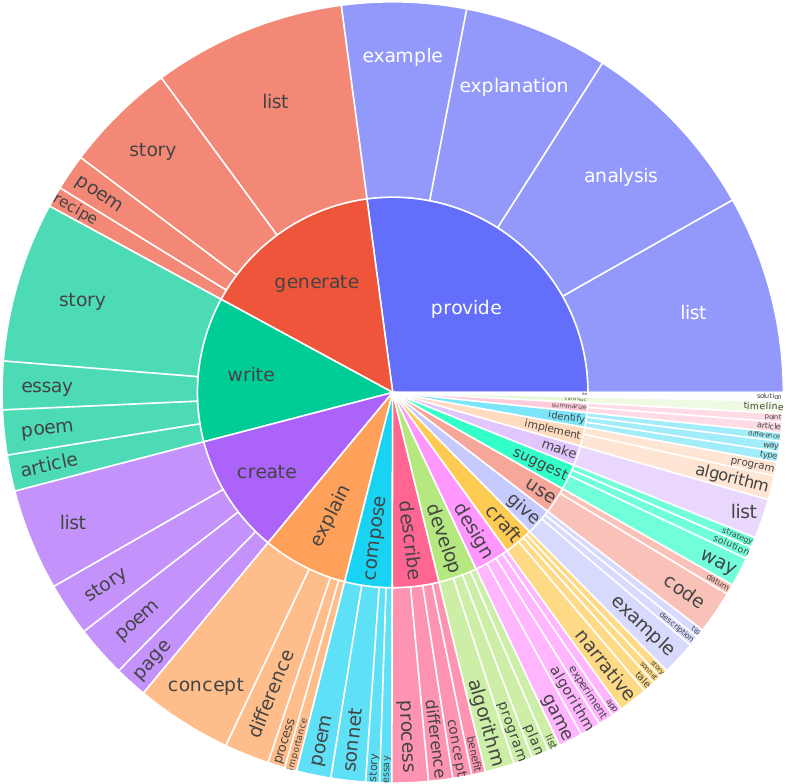}
    \caption{High Instruction Following Difficulty.}

  \end{subfigure}%
  \hspace{0.01\textwidth}%
  \begin{subfigure}[b]{0.325\textwidth}
    \centering
    \includegraphics[width=\textwidth]{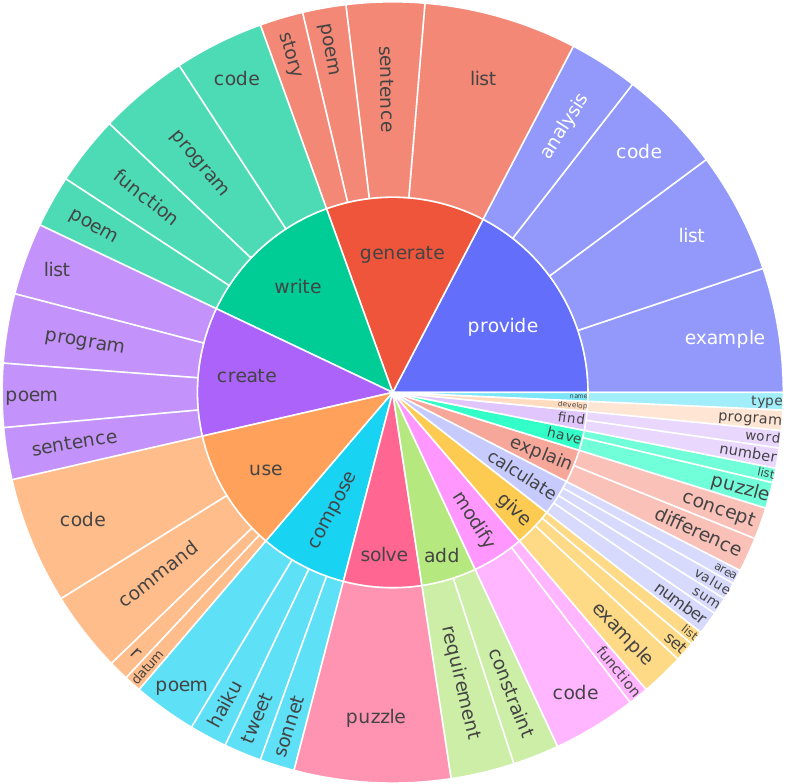}
    \caption{High In-context Influence.}

  \end{subfigure}%
  \hspace{0.01\textwidth}%
  \begin{subfigure}[b]{0.325\textwidth}
    \centering
    \includegraphics[width=\textwidth]{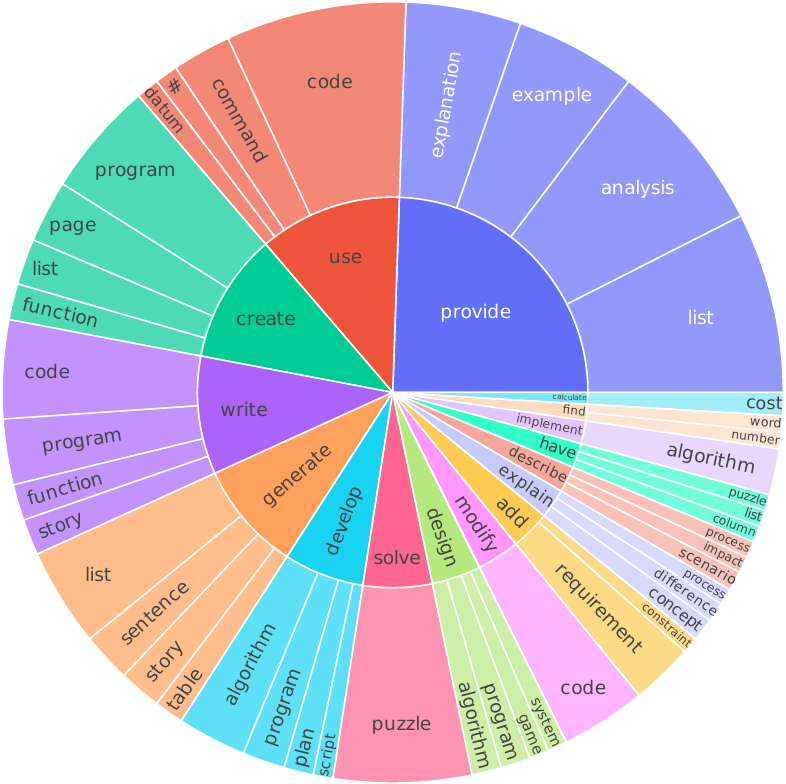}
    \caption{Random.}

  \end{subfigure}

  \caption{Visualization of the verb–noun structures in instructions selected by the three data-selection strategies on Wizard. The inner ring shows the predominant verbs, the outer ring displays the nouns that co-occur directly with those verbs.}
  \label{fig:cake3}
\end{figure*}

\newpage

\section{Additional Results}

\subsection{Additional Benchmark: IFEval}

To further assess whether selected instruction-tuning data improves instruction-following ability beyond general knowledge and reasoning benchmarks, we additionally evaluate all methods on IFEval~\citep{zhou2023instruction}. IFEval measures how well language models follow \emph{verifiable} instructions, such as output format constraints, keyword requirements, and length constraints, and thus provides a more targeted test of instruction following than broad capability benchmarks. Following the standard evaluation protocol, we report the average over four metrics: prompt-level strict (Pr(S)), prompt-level loose (Pr(L)), instruction-level strict (Ins(S)), and instruction-level loose (Ins(L)). As shown in Table~\ref{tab:ifeval}, IFEval complements our main benchmarks by directly evaluating whether models can precisely comply with explicit user requirements.

The results reveal a clear pattern: in all four settings, the model trained on the full dataset achieves the best performance. Although data selection methods improve performance on knowledge- and answer-quality-oriented benchmarks, these gains do not carry over to stronger instruction-following ability on IFEval. 
This suggests that instruction following depends more on sufficient data coverage, while being relatively less sensitive to data quality than other capabilities. 
Among the selection-based methods, our method remains competitive and achieves the best result in the Llama3.1-8B / Alpaca-GPT4 and Mistral-7B / Alpaca-GPT4 settings. Still, the consistent advantage of the full-data baseline indicates that, for instruction-following evaluation, maintaining training scale is more important than selecting a smaller but higher-quality subset. 
Overall, these findings complement our main results by showing that data selection is especially beneficial for knowledge-intensive and response-quality benchmarks, whereas instruction-following ability appears to benefit more from scale.


\subsection{Additional Data Visualization}
We provide root verbs and their direct objects from our
extracted instruction subset of Wizard70k in Figure~\ref{fig:cake3}.

\end{document}